\documentclass[letterpaper, 10 pt, conference]{ieeeconf}  

\IEEEoverridecommandlockouts                              

\usepackage{hyperref}
\usepackage{graphicx}
\usepackage{caption}
\usepackage[noadjust]{cite}
\usepackage[nointegrals]{wasysym}

\graphicspath{ {images/} }
\usepackage{xcolor}
\usepackage{booktabs}
\usepackage{tabularx}
\usepackage{multirow}
\usepackage{rotating}
\newcolumntype{C}{>{\centering\arraybackslash}X}
\newcolumntype{T}{>{\hsize=20pt}X}
\newcolumntype{Y}{>{\hsize=10pt}C}
\newcolumntype{S}{>{\hsize=0.15\hsize}C}
\newcolumntype{B}{>{\hsize=0.70\hsize}C}
\newcolumntype{A}{>{\hsize=0.15\hsize}C}

\usepackage{algorithm}
\usepackage{algpseudocode}
\usepackage{algorithmicx}

\algrenewcommand{\algorithmiccomment}[1]{{\scriptsize{\textcolor{blue} {\ttfamily// #1}}}}
\algdef{SE}[DOWHILE]{Do}{DoWhile}{\algorithmicdo}[1]{\algorithmicwhile\ #1}%

\usepackage{amsmath}
\usepackage{amssymb}

\usepackage{newfloat}
\usepackage{listings}
\floatstyle{ruled}
\newfloat{listing}{tb}{lst}{}
\floatname{listing}{Listing}

\title{\LARGE \bf
Learning Type-Generalized Actions for Symbolic Planning
}

\author{Daniel Tanneberg\textsuperscript{\includegraphics[scale=0.08]{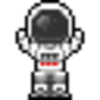}} and Michael Gienger\textsuperscript{\includegraphics[scale=0.08]{asimo.png}}
\thanks{\includegraphics[scale=0.08]{asimo.png} Honda Research Institute EU, Germany
        {\tt\small \{daniel.tanneberg, michael.gienger\}@honda-ri.de}}%
}

\newcommand{\plotlearningoverview}{
	\begin{figure*}[h]
		\centering
		\includegraphics[width=0.95\textwidth]{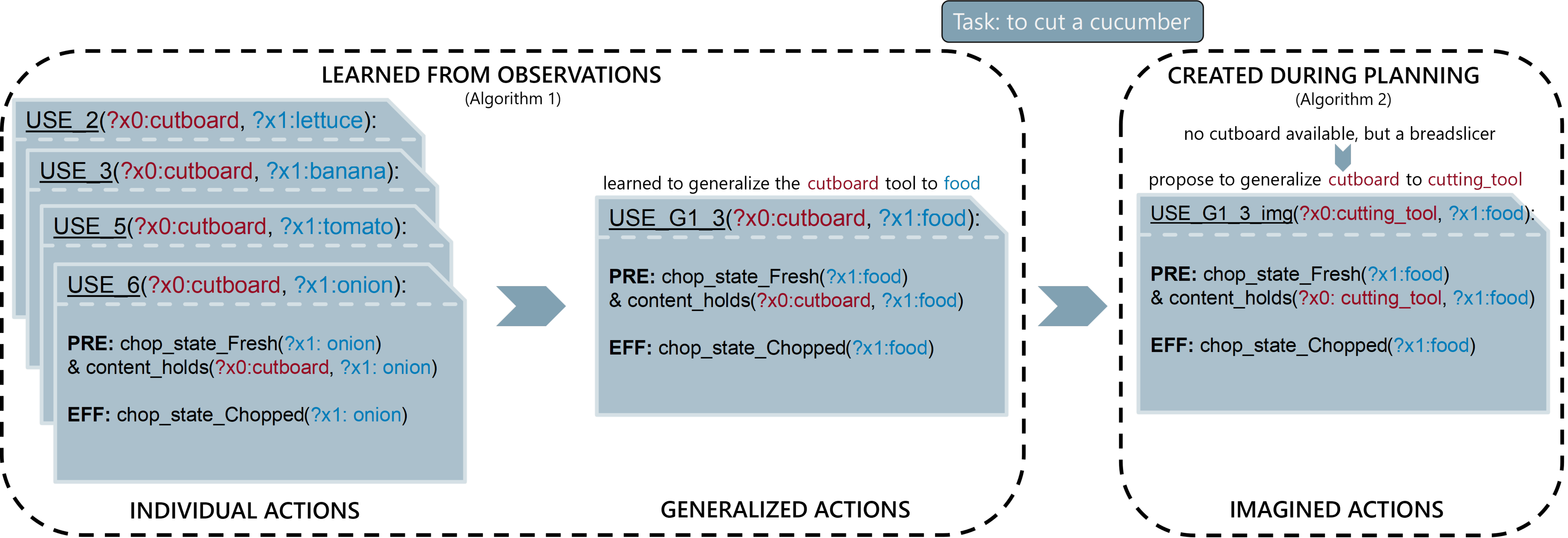}
		\caption{Simplified sketch of the action learning and creating algorithms, showcased on learning to chop different foods.
			First, Algorithm~\ref{alg:learning} learns a set of individual actions from observations, i.e., individual actions for cutting \textit{lettuce}, \textit{banana}, \textit{tomato}, and \textit{onion} with the \textit{cutboard}.
			Next, these individual actions are generalized based on their effects and the entity hierarchy to create a generalized cutting action for \textit{food}.
			During planning, Algorithm~\ref{alg:imagine} can create imagined generalized actions, i.e., proposed generalizations based on effect, entity hierarchy, and available entities in the current planning problem.
			Note, action preconditions (PRE) and effects (EFF) show a simplified subset for visualization and explainability.
			Shown are the learned lifted actions, where \texttt{?x0} is the variable name followed by its type.
		}
		\label{fig:learning_overview}
	\end{figure*}
}

\newcommand{\plotovercooked}{
	\begin{figure}[!t]
		\centering
		\includegraphics[width=0.23\textwidth]{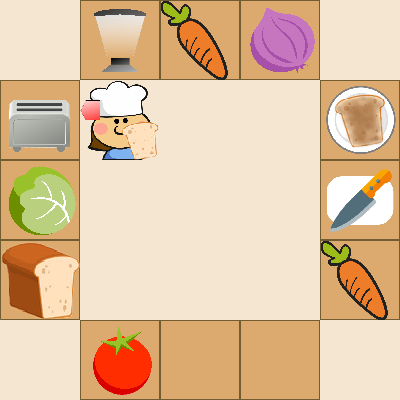}
		\caption{Screenshot of the kitchen scenario, showing the agent and different entities (food and tools) to interact with.}
		\label{fig:kitchen}
	\end{figure}
}

\newcommand{\plothierarchy}{
	\begin{figure}[!t]
		\centering
		\includegraphics[width=0.48\textwidth]{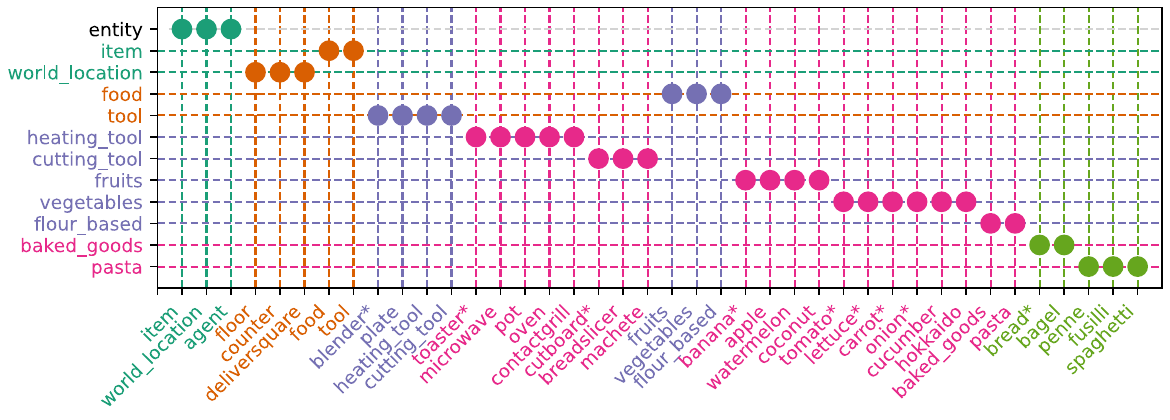}
		\caption{Visualization of the used entity hierarchy $\mathcal{H}$, with \textit{parents} on the y-axis and \textit{children} on the x-axis.
			A dot indicates a connection, the color indicates the depth levels inside the hierarchy (top to bottom), and * indicates the entity was seen in the training data.}
		\label{fig:hierarchy}
	\end{figure}
}

\newcommand{\plotresultsplit}{
	\begin{figure}[!t]
		\centering
		\includegraphics[width=0.5\textwidth]{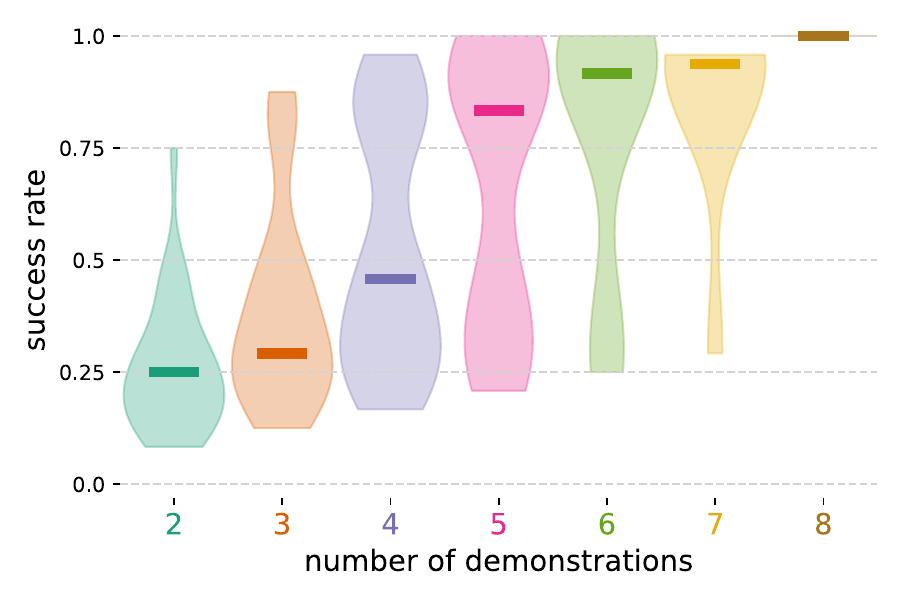}
		\caption{Fraction of tasks where a successful plan was found per number of demonstrations used for learning.
		The shaded violion shows the success ratio distribution of the learned action sets, and the colored bar marks the median.}
		\label{fig:result_summary}
	\end{figure}
}

\newcommand{\plotalgorithmlearning}{
	\begin{algorithm}[!t]
		\caption{Learning generalized actions}
		\label{alg:learning}
		\begin{algorithmic}[1] 
			\Require Training data $\mathcal{D}$, entity hierarchy $\mathcal{H}$
			\Ensure Learned action set $\mathcal{A}$
			\Statex \Comment{Learn individual actions}
			\Statex \Comment{cluster data by action $a_d$ and lifted effect $\mathit{eff}$}
			\State clusters$[\mathit{(a_d,eff)}]$ $\gets$ \Call{ClusterLiftedEffects}{$\mathcal{D}$}
			\Statex \Comment{extract action for each cluster}
			\For{\textbf{each} $\mathcal{D}_{\mathit{a_d,eff}} \in \textrm{clusters}$}
			\Statex \hspace{1em} \Comment{extract preconditions for each effect}
			\State $\mathit{pre}$ $\gets$ \Call{ExtractPreconditions}{$\mathcal{D}_{\mathit{a_d,eff}}, \mathcal{D}_{\mathit{a_d,\cdot}}$}
			\Statex \hspace{1em} \Comment{create action and add it to the action set}
			\State $a$ $\gets$ \Call{CreateAction}{$a_d, \mathit{pre}, \mathit{eff}$}
			\State $\mathcal{A}$ $\gets$ $\mathcal{A} \cup \{a\}$
			\EndFor
			\Statex \Comment{Iteratively generalize actions}
			\Do
			\State $\mathcal{A}_{\mathit{old}} \gets \mathcal{A}$
			\For{\textbf{each} $\{(a_i, a_j) \ | \ a_i \in \mathcal{A}, a_j \in \mathcal{A}, a_i \neq a_j \}$}
			\Statex \hspace{2.em} \Comment{consider actions with same effect predicates}
			\If{$a_i^{\mathit{eff}} = a_j^{\mathit{eff}}$}
			\Statex \hspace{4.5em} \Comment{create generalized parameters}
			\State $\mathit{p_g} \gets$ \Call{CreateGenPars}{$a_i, a_j$}
			\Statex \hspace{4.5em} \Comment{create and score generalized action}
			\State $a_g, v_g \gets$ \Call{CreateGenAction}{$a_i, a_j, p_g$}
			\Statex \hspace{4.5em} \Comment{score individual actions}
			\State $v_i \gets$ \Call{ScoreAction}{$a_i$}
			\State $v_j \gets$ \Call{ScoreAction}{$a_j$}
			\Statex \hspace{4.5em} \Comment{compare scores \& keep new action}
			\If{$v_g \geq (v_i + v_j)/2$}
			\State $\mathcal{A} \gets (\mathcal{A} \backslash \{a_i, a_j\}) \cup \{a_g\}$
			\EndIf
			\EndIf
			\EndFor
			\DoWhile{$\mathcal{A}_{\mathit{old}} \neq \mathcal{A}$}
		\end{algorithmic}
	\end{algorithm} 
}

\newcommand{\plotalgorithmimagine}{
	\begin{algorithm}[!t]
		\caption{Imagine generalized actions}
		\label{alg:imagine}
		\begin{algorithmic}[0] 
			\Require Action set $\mathcal{A}$, grounded action set $\mathcal{A}_g$, entity hierarchy $\mathcal{H}$, task $\mathcal{T}$
			\Ensure Enhanced action set $\mathcal{A}$
			\Statex \Comment{find unreachable goal predicates with initial state \& grounded actions}
			\State $\mathcal{U} \gets$ \Call{GetUnreachableGoals}{$\mathcal{T}_{\mathit{init}}, \mathcal{A}_g$}
			\While{$|\mathcal{U}| > 0$}
			\Statex \hspace{1.5em} \Comment{get potential actions for unreached goal}
			\State $u_g \gets \mathcal{U}.pop()$
			\State $\mathcal{A}_p \gets$ \Call{GetPotentialActions}{$\mathcal{A}, u_g$}
			\Statex \hspace{0.6em} \Comment{generalize actions \& save parameter substitutions}
			\For{\textbf{each} $a_p \in \mathcal{A}_p$}
			\State $a_i, p_i \gets$ \Call{CreateImaginedAction}{$a_p, \mathcal{T}$}
			\State $\mathcal{A} \gets \mathcal{A} \cup \{a_i\}$
			\State $\mathcal{P} \gets \mathcal{P} \cup \{p_i\}$
			\Statex \hspace{2em} \Comment{add unreached preconditions to unreached set}
			\State $\mathcal{U} \gets \mathcal{U} \ \cup$ \Call{GetUnreachablePreconds}{$a_i$}
			\EndFor
			\EndWhile
			\Statex \Comment{transfer parameter substitutions on actions with same parameters}
			\For{\textbf{each} $p_i \in \mathcal{P}$}
			\For{\textbf{each} $a \in \mathcal{A} \mid p_i \in a^{\mathit{params}}$}
			\State $a_p \gets$ \Call{CreateSubstitutedAction}{$a, p_i$}	
			\State $\mathcal{A} \gets \mathcal{A} \cup \{a_p\}$
			\EndFor
			\EndFor
		\end{algorithmic}
	\end{algorithm}
}

\newcommand{\plotresulttable}{
	\begin{table*}[!t]
		\centering
		\footnotesize
		\begin{tabularx}{0.99\textwidth}{@{}YT*{4}{!{\vrule width 1.2pt}S|B|A}}
			& & \multicolumn{3}{c!{\vrule width 1.2pt}}{\textbf{generalized+imagined}} & \multicolumn{3}{c!{\vrule width 1.2pt}}{\textbf{generalized}} & \multicolumn{3}{c}{\textbf{individual+imagined}}  & \multicolumn{3}{c}{\textbf{individual}} \\
			& & $\#$p & time (s) & $|p|$ & $\#$p & time (s) & $|p|$ & $\#$p & time (s) & $|p|$ & $\#$p & time (s) & $|p|$ \\
			\midrule[1.2pt]
			\multirow[c]{4}{*}[-8pt]{\begin{sideways}\textbf{S1}\end{sideways}} 
			& T1.1 & 
			1 & 1.12 $\pm$ 0.03 & 5 & 
			1 & 1.12 $\pm$ 0.03 & 5 & 
			1 & 0.30 $\pm$ 0.02 & 5 &
			1 & 0.31 $\pm$ 0.03 & 5 \\
			\cmidrule{2-14}
			& T1.2 & 
			1 & 1.08 $\pm$ 0.02 & 5 & 
			1 & 1.07 $\pm$ 0.03 & 5 & 
			1 & 0.32 $\pm$ 0.03 & 5 &
			1 & 0.32 $\pm$ 0.03 & 5 \\
			\cmidrule{2-14}
			& T1.3 & 
			1 & 1.27 $\pm$ 0.03 & 15 & 
			1 & 1.28 $\pm$ 0.05 & 15 & 
			1 & 0.62 $\pm$ 0.03 & 15 &
			1 & 0.61 $\pm$ 0.03 & 15 \\
			\cmidrule{2-14}
			& T1.4 & 
			1 & 1.30 $\pm$ 0.08 & 10 & 
			1 & 1.22 $\pm$ 0.05 & 10 & 
			1 & 0.38 $\pm$ 0.03 & 10 &
			1 & 0.39 $\pm$ 0.03 & 10 \\
			\midrule[1.2pt]
			\multirow[c]{4}{*}[-8pt]{\begin{sideways}\textbf{S2}\end{sideways}} 
			& T2.1 & 
			1 & 1.16 $\pm$ 0.03 & 5 & 
			1 & 1.11 $\pm$ 0.03 & 5 & 
			1 & 0.51 $\pm$ 0.03 & 8 &
			1 & 0.52 $\pm$ 0.04 & 8 \\
			\cmidrule{2-14}
			& T2.2 & 
			1 & 1.27 $\pm$ 0.05 & 10 & 
			1 & 1.24 $\pm$ 0.03 & 10 & 
			- & - & - &
			- & - & - \\
			\cmidrule{2-14}
			& T2.3 & 
			1 & 1.36 $\pm$ 0.02 & 13 & 
			1 & 1.33 $\pm$ 0.03 & 13 & 
			- & - & - &
			- & - & - \\
			\cmidrule{2-14}
			& T2.4 & 
			1 & 1.25 $\pm$ 0.03 & 13 & 
			1 & 1.26 $\pm$ 0.02 & 13 & 
			1 & 0.56 $\pm$ 0.03 & 16 &
			1 & 0.58 $\pm$ 0.04 & 16 \\
			\midrule[1.2pt]
			\multirow[c]{4}{*}[-8pt]{\begin{sideways}\textbf{S3}\end{sideways}} 
			& T3.1 & 
			1 & 2.88 $\pm$ 0.08 & 15 & 
			1 & 2.89 $\pm$ 0.05 & 15 & 
			- & - & - &
			- & - & - \\
			\cmidrule{2-14}
			& T3.2 & 
			1 & 2.52 $\pm$ 0.06 & 18 & 
			1 & 2.49 $\pm$ 0.04 & 18 & 
			- & - & - &
			- & - & - \\
			\cmidrule{2-14}
			& T3.3 & 
			1 & 6.73 $\pm$ 0.08 & 25 & 
			1 & 6.73 $\pm$ 0.08 & 25 & 
			- & - & - &
			- & - & - \\
			\cmidrule{2-14}
			& T3.4 & 
			1 & 7.00 $\pm$ 0.03 & 30 & 
			1 & 6.98 $\pm$ 0.04 & 30 & 
			- & - & - &
			- & - & - \\
			\midrule[1.2pt]
			\multirow[c]{4}{*}[-8pt]{\begin{sideways}\textbf{S4}\end{sideways}} 
			& T4.1 & 
			1 & 1.67 $\pm$ 0.02 & 5 & 
			- & - & - & 
			1 & 3.74 $\pm$ 0.03 & 5 &
			- & - & - \\
			\cmidrule{2-14}
			& T4.2 & 
			1 & 1.87 $\pm$ 0.02 & 10 & 
			- & - & - & 
			1 & 4.08 $\pm$ 0.06 & 10 &
			- & - & - \\
			\cmidrule{2-14}
			& T4.3 & 
			1 & 1.93 $\pm$ 0.04 & 13 & 
			- & - & - & 
			8 & 66.92 $\pm$ 1.54 & 15 &
			- & - & - \\
			\cmidrule{2-14}
			& T4.4 & 
			1 & 1.92 $\pm$ 0.04 & 16 & 
			- & - & - & 
			- & - & - &
			- & - & - \\
			\midrule[1.2pt]
			\multirow[c]{4}{*}[-8pt]{\begin{sideways}\textbf{S5}\end{sideways}} 
			& T5.1 & 
			1 & 1.11 $\pm$ 0.04 & 5 & 
			1 & 1.12 $\pm$ 0.03 & 5 & 
			2 & 7.84 $\pm$ 0.10 & 5 &
			- & - & - \\
			\cmidrule{2-14}
			& T5.2 & 
			1 & 1.39 $\pm$ 0.01 & 10 & 
			1 & 1.37 $\pm$ 0.04 & 10 & 
			2 & 6.30 $\pm$ 0.12 & 10 &
			- & - & - \\
			\cmidrule{2-14}
			& T5.3 & 
			1 & 1.30 $\pm$ 0.04 & 10 & 
			1 & 1.31 $\pm$ 0.04 & 10 & 
			9 & 44.12 $\pm$ 0.64 & 10 &
			- & - & - \\
			\cmidrule{2-14}
			& T5.4 & 
			1 & 1.35 $\pm$ 0.05 & 13 & 
			1 & 1.33 $\pm$ 0.04 & 13 & 
			9 & 46.45 $\pm$ 0.40 & 15 &
			- & - & - \\
			\midrule[1.2pt]
			\multirow[c]{4}{*}[-8pt]{\begin{sideways}\textbf{S6}\end{sideways}} 
			& T6.1 & 
			1 & 2.69 $\pm$ 0.07 & 18 & 
			- & - & - & 
			- & - & - &
			- & - & - \\
			\cmidrule{2-14}
			& T6.2 & 
			1 & 4.84 $\pm$ 0.09 & 23 & 
			- & - & - & 
			- & - & - &
			- & - & - \\
			\cmidrule{2-14}
			& T6.3 & 
			1 & 18.74 $\pm$ 0.26 & 28 & 
			- & - & - & 
			- & - & - &
			- & - & - \\
			\cmidrule{2-14}
			& T6.4 & 
			1 & 12.55 $\pm$ 0.24 & 36 & 
			- & - & - & 
			- & - & - &
			- & - & - \\
			\midrule[1.2pt]
		\end{tabularx}
		\caption{Evaluation of learned actions on different sets \textbf{S}\# of planning tasks \textbf{T}\#.x (see \textit{Results} for task sets definition). 
			Comparing the results of the baseline learning \textit{individual} actions and the proposed Algorithm~\ref{alg:learning} for learning \textit{generalized} actions.
			Both methods can be extended with the second proposed Algorithm~\ref{alg:imagine} (\textit{+imagined}).
			Results obtained over $5$ runs each with $120$s timeout for planning and randomized object placements, while $-$ indicates no successful plan was found.
			$\#p$ show how many plans where proposed until successful execution.
			Time measures the total planning time in \textit{s}, where mean and standard deviation are reported.
			$|p|$ is an indicator for the task complexity by measuring the length of the successful plan as the length of the action sequence to solve it.
			The environments always consist of $32$ objects.
		}
		\label{tab:planning_results}
		\vspace{-0pt}
	\end{table*}
}

\usepackage{fancyhdr}
\fancypagestyle{firstpage}{
	\fancyhf{}
	\fancyfoot[L]{\scriptsize \copyright 2023 IEEE. Personal use of this material is permitted. Permission from IEEE must be obtained for all other uses, in any current or future media, including reprinting/republishing this material for advertising or promotional purposes, creating new collective works, for resale or redistribution to servers or lists, or reuse of any copyrighted component of this work in other works.
		DOI: \href{https://doi.org/10.1109/IROS55552.2023.10342301}{10.1109/IROS55552.2023.10342301}}

}

\begin{document}

\maketitle
\thispagestyle{firstpage}

\begin{abstract}
Symbolic planning is a powerful technique to solve complex tasks that require long sequences of actions and can equip an intelligent agent with complex behavior.
The downside of this approach is the necessity for suitable symbolic representations describing the state of the environment as well as the actions that can change it.
Traditionally such representations are carefully hand-designed by experts for distinct problem domains, which limits their transferability to different problems and environment complexities.
In this paper, we propose a novel concept to generalize symbolic actions using a given entity hierarchy and observed similar behavior.
In a simulated grid-based kitchen environment, we show that type-generalized actions can be learned from few observations and generalize to novel situations.
Incorporating an additional on-the-fly generalization mechanism during planning, unseen task combinations, involving longer sequences, novel entities and unexpected environment behavior, can be solved.
\end{abstract}

\section{Introduction}
In order for embodied intelligent agents like robots to solve complex, longer horizon tasks, they need to be able to plan ahead.
They need to have some kind of abstract model of the environment, their abilities and strategies, and the task to be solved~\cite{tenenbaum2011grow,lake2017building,Konidaris2019,tanneberg2020evolutionary}.
This idea is the core of (classical) artificial intelligence planning, or symbolic planning, and has a long research history~\cite{Arora2018}.
In many robotic systems and applications using symbolic planning, the required representations, of states and actions, are domain-specific and hand-designed by domain experts.

In recent years more and more research has tackled the problem of learning the required symbolic representations, especially in the robot learning, and task and motion planning (TAMP) communities, where hand-designed representations are often used~\cite{zhu2021hierarchical,Garrett2021}.
Many approaches can be divided into two (non-exclusive) broad categories, the ones focusing on learning the action representation assuming a known state representation~\cite{pasula2007learning,aineto2019learning,Mitrevski2020,Kim2020,lamanna2021online,Silver2021,suarez2021online}, e.g., given as known logical predicates.
Another line of research tackles the problem of learning a state representation, either additionally to the actions or separately, from raw (sensor) input~\cite{Kulick2013,Ahmadzadeh2015,Aksoy2016,Asai2018,Konidaris2018,yuan2021sornet,Chitnis2021,Cingillioglu2021,rodriguez2021learning}.

In this work, we focus on learning symbolic planning actions from few observations with known state representation and leveraging entity hierarchies.
More precise, we learn type-generalized actions that can transfer to a variety of novel situations and entities, instead of object-type specific actions~\cite{Abdo2013,Ahmadzadeh2015,Asai2018,aineto2019learning,Silver2021,Chitnis2021,diehl2021automated}.
Type-generalization is a powerful concept that allows a straightforward transfer to unknown situations and entities, that is based on similar behavior of entities.
In addition to learning these type-generalized actions from few observations based on a given entity hierarchy and observed similar behavior, we also propose a method integrated in a standard search based heuristic planner that \textit{imagines} new generalized action proposals during planning.
These learned and imagined lifted type-generalized actions can be grounded with available objects in a concrete planning problem.
An illustrative overview of the framework is shown in Figure~\ref{fig:learning_overview}. 
In a simulated kitchen environment (Figure~\ref{fig:kitchen}), we show the benefits of such type-generalized actions in various generalization complexities.

\plotlearningoverview

\section{Learning and Imagine Generalized Actions}
In this section, we first describe the proposed approach for learning generalized actions (summarized in Algorithm~\ref{alg:learning}) from observations, and then the proposed approach to create imagined generalized actions on demand during planning (summarized in Algorithm~\ref{alg:imagine}).
An illustrative example for both algorithms is given in Figure~\ref{fig:learning_overview}.
When speaking of actions, we refer to symbolic descriptions of actions an agent can perform in an environment and which can be used for symbolic planning, i.e., that consist of parameters, preconditions that must be fulfilled to execute an action, and effects the executed action implies.
Here, parameters are typed placeholders for objects that are used in the preconditions and effects.
The preconditions are lifted logical predicates that must be fulfilled in the current state in order to execute an action.
The effect describes the lifted logical predicates that change through the execution of the action -- a common description using the Planning Domain Definition Language (PDDL)~\cite{fox2003pddl2}.
The actions can be grounded by assigning the placeholder parameters (lifted) specific objects in the current environment and passing these substitutions to the preconditions and effects -- e.g., a lifted predicate like 

\texttt{content\_holds(?x0:cutboard, ?x1:food)} 
\\
describing that an object \texttt{?x1} of type \texttt{food} needs to be the content of an object \texttt{?x0} of type \texttt{cutboard} can be grounded in a given environment with all suitable objects into
\begin{footnotesize}
\texttt{content\_holds(cutboard\_1:cutboard, tomato\_1:tomato)},\\ 
\texttt{content\_holds(cutboard\_2:cutboard, onion\_1:onion)}, \\
\texttt{content\_holds(cutboard\_1:cutboard, onion\_2:onion)}, 
\end{footnotesize}
\\
and so on. 
While the observations in the training data are grounded observations (set of grounded predicates), the learned actions are lifted to transfer to other object instances.

\subsection{Learning Generalized Actions}
Learning is done from observations, where each observation consists of a sequence of $(s,a_d,s')$ tuples -- where $s$ is the state before action $a_d$ was executed, and $s'$ the state after the execution -- creating the training data $\mathcal{D}$.
The action description $a_d$ consists of the action's name and the entities it interacts with, and $s$ is a set of grounded logical predicates as described before.
Given the training data $\mathcal{D}$ and an entity hierarchy $\mathcal{H}$ (e.g., see Figure~\ref{fig:hierarchy}), we want to learn a set of actions $\mathcal{A}$.

The proposed approach (Algorithm~\ref{alg:learning}) works in two phases -- first, \textit{individual} actions are learned from the training data, and second, learned actions are iteratively generalized until convergence.
The learning of individual actions is based on the recent LOFT algorithm~\cite{Silver2021}.
We explain our approach following the pseudocode given in Algorithm~\ref{alg:learning}.

The first phase consists of three main steps, which are described in detail next, and learns the individual actions from the given training data.

\textit{ClusterLiftedEffects:} First, for each tuple $(s,a_d,s')$ in $\mathcal{D}$ the grounded effect is calculated by the difference between $s$ and $s'$.
These observation tuples are then clustered by the action description $a_d$ and their effects, where two observation tuples of the same action belong to the same cluster if their effects can be \textit{unified} -- meaning, there exists a bijective mapping between the objects in the two observation tuples such that the effects are equivalent up to this mapping, i.e., they have a similar effect.
The observations in the effect clusters are then lifted, i.e., replacing the concrete object with variables.

\textit{ExtractPreconditions:} Next, for each effect cluster the preconditions are extracted. 
While the LOFT algorithm uses an inner and outer search to learn the preconditions, we adapted a simpler and faster approach to extract the preconditions.
The preconditions are calculated as the intersection of all states $s$ of the observations in the cluster, i.e., the predicates that are true in all observations before action execution.
This approach tends to produce over-specialized preconditions, but in the generalization step later on, preconditions are potentially further relaxed by looking at multiple actions at once and, hence, more data.
Moreover, we consider learning from (human) observations, meaning very little data and especially no negative examples (see Experiments section), making the learning of precise preconditions even harder.
Hence, we use the proposed approach in this step for simplicity and efficiency, and leverage additional data during the generalization to update the preconditions (and in a future step during interaction and feedback).

\textit{CreateAction:} Lastly, we create a learned action $a$ with the action description $a_d$, the extracted preconditions \textit{pre}, the calculated effect \textit{eff}, and add it to the action set $\mathcal{A}$.

The second phase iteratively tries to find actions with similar effects and uses the given entity hierarchy information to generalize those.
For each pair of actions $(a_i, a_j)$ with similar effects, where similar effects are defined as having the same set of predicates in the effect set $a_i^{\mathit{eff}} = a_j^{\mathit{eff}}$, possible variable type generalizations are generated and tested.

\textit{CreateGenPars:} Calculates the generalized parameters $p_g$ from the actions $(a_i, a_j)$ by using the given entity hierarchy.
For each corresponding parameter pair from $a_i$ and $a_j$, the lowest common ancestor (LCA) is calculated and added as proposed generalization to $p_g$.
The LCA is used for generalization as we want to find the \textit{most specific} generalization.
With a broader -- higher ancestor in the hierarchy -- generalization of the lifted action, the more grounded actions need to be considered during planning, which slows down planning drastically.

\textit{CreateGenAction:} Creates the generalized action $a_g$ by replacing the action parameters with the proposed new parameters $p_g$.
Furthermore, it updates the learned preconditions by considering the intersection of the preconditions of $a_i$ and $a_j$ (after parameter replacement) together with all elements of the powerset ($\mathbb{P}(\cdot)$) of the difference of their preconditions, i.e., creates a set of candidate preconditions
\begin{align}
	\{ (a_i^{\mathit{pre}} \cap  a_j^{\mathit{pre}}) \cup \mathit{pre} \ &| \ \mathit{pre} \in  \mathbb{P}\big( d \big)  \} \ , \nonumber \\
	\text{with} \ d &= (a_i^{\mathit{pre}} \cup  a_j^{\mathit{pre}}) - (a_i^{\mathit{pre}} \cap  a_j^{\mathit{pre}}) \ . \nonumber
\end{align}
The generalized action $a_g$ is created with the best performing preconditions from the candidate set with the score $v_g$.
The score $v_g$ is calculated as the Recall using all observation clusters from $a_i$ and $a_j$ as positive data.

\textit{ScoreAction:} Scores the individual action $s_i$ and $s_j$ similarly as described before on their respective observation clusters to determine if the generalized action $s_g$ should be kept.
The generalized action $s_g$ replaces the individual actions $s_i$ and $s_j$ if its performance is at least as good as the average performance of $s_i$ and $s_j$.

This approach iterates until convergence, meaning the action set $\mathcal{A}$ does not change anymore.
Note, in the above description $a_i$ and $a_j$ are always referred to as individual actions for explainability, but due to the iterations they can be previously generalized actions already.

\plotalgorithmlearning

\subsection{Imagine Generalized Actions}
While learning generalized actions from observations as described before can transfer to novel tasks and objects, it is limited to generalizations of observed behavior.
For example, considering the entity hierarchy in the experiments (Figure~\ref{fig:hierarchy}), if only chopping of different \textit{vegetables} is shown, the learned generalized action will be applicable to known and novel vegetables, but cannot be grounded with \textit{fruits}.
Or differently, if chopping is only observed with the \textit{cutboard}, a novel chopping tool like the \textit{breadslicer} cannot be grounded.

To cope with such missing generalizations, and also with over-generalizations (e.g., trying to chop a \textit{coconut} with the \textit{cutboard}) or unexpected behavior during execution (e.g., an object does not behave as observed anymore because it is broken or similar), we propose a second action generalization algorithm that works \textit{on-the-fly} during planning.
The approach is summarized in Algorithm~\ref{alg:imagine}.
We refer to the resulting actions as \textit{imagined} generalized actions as the agent \textit{imagines} a new generalization without observed data, and these actions need to be validated by executing the found plan, or by proposing it to a human for feedback.

Given a (lifted) action set $\mathcal{A}$, its grounded action set $\mathcal{A}_g$, an entity hierarchy $\mathcal{H}$, and a planning task $\mathcal{T}$ (including the initial state $\mathcal{T}_{\mathit{init}}$), the proposed imagined generalization approach returns an enhanced action set $\mathcal{A}$.

\textit{GetUnreachableGoals:} First, using the initial state $\mathcal{T}_{\mathit{init}}$ and the grounded action set $\mathcal{A}_g$, unreachable goal predicates $\mathcal{U}$ are calculated, i.e., predicates in the planning task goal description that cannot be reached with $\mathcal{A}_g$ starting from $\mathcal{T}_{\mathit{init}}$.
If all goal predicates can be reached, no imagined generalized actions are necessary.

\textit{GetPotentialActions:}
Iteratively for each unreachable goal predicate $u_g \in \mathcal{U}$ potentially suitable actions for generalization are calculated from $\mathcal{A}$.
That are those actions with the unreachable goal predicate in their effect set.

\textit{CreateImaginedAction:} For each potential action the generalization is then calculated by finding the lowest common ancestor of the variables in the unreachable goal predicate and the matching effect predicate of the potential action.
Additionally, all variables of the potential action are checked and generalized to the lowest common ancestor of itself and the available objects in the planning task.
With this parameter generalization $p_i$, a new action $a_i$ is created and added to the action set $\mathcal{A}$.

\textit{GetUnreachablePreconds:} Checks if the predicates in the preconditions of the new action $a_i$ can be reached and adds unreachable predicates to the set of unreachable goal predicates $\mathcal{U}$.

\textit{CreateSubstitutedAction:} Propagates the proposed parameter generalizations $p_i$ to all actions that have the same set of parameters as the original potential action, and adds those generalized actions to the enhanced action set $\mathcal{A}$ as well.

After this whole process, the action set $\mathcal{A}$ contains new generalized action proposals that can be used to solve the planning problem, that was not solvable with the original action set.

\plotalgorithmimagine

\section{Planning with Generalized Actions}
\label{sec:planning}
In order to solve planning tasks with the learned generalized actions, which are straightforwardly mapped into PDDL representations, we use the planning as heuristic search approach~\cite{bonet2001planning} with an A* search algorithm~\cite{Hart1968} and adapted ordered landmarks heuristic~\cite{hoffmann2004ordered,richter2010landmark}.
The planner returns the first found plan greedily, favoring planning speed, but can be triggered to continue planning if another plan is required, or if there is planning time budget left to improve the plan.
We used this simple implementation to have full control in the evaluations and as the planning approach itself is not the focus of this paper.
A landmark is a grounded predicate that needs to be fulfilled (reached) during planning to solve the task.
We cluster the ordered landmarks based on the objects in the planning goal description, resulting in multiple landmark sequences, allowing the heuristic planner to explore with less restrictions.
The heuristic is based on the idea to count the unreached landmarks~\cite{richter2010landmark}, where a landmark is only marked as reached if all predecessor landmarks have been reached before.
We add penalties to the heuristic for actions that do not operate on goal objects (objects that are defined in the planning goal) and that do not reach new landmarks.

Given a planning task $\mathcal{T}$ -- consisting of the initial state of the environment and the goal description -- and the action set $\mathcal{A}$, the planner creates the grounded action set $\mathcal{A}_g$ from $\mathcal{A}$, the entity hierarchy, and $\mathcal{T}_{\mathit{init}}$.
Then Algorithm~\ref{alg:imagine} is called to check for potential imagined action generalizations -- if $|\mathcal{U}| = 0$, i.e., the task can be solved with the current $\mathcal{A}_g$, no imagined actions are added.
When a plan is found, the agent tries to execute it in the environment using a fixed deterministic policy, where after each executed action, the agent checks if the expected effect (the effect set of the executed action) matches the observed effect (the change in the environment).
If there is mismatch, the execution is aborted and the planner searches for a new plan proposal.
The specific objects involved in the aborting action are excluded during this search continuation and Algorithm~\ref{alg:imagine} is triggered again to check if new imagined actions should be added due to this.
This simple strategy to exclude such objects assumes that the excluded object is itself the source of the unexpected effect (e.g., a tool is broken and not working anymore, or a human gives feedback) and guides the planner to search for alternative objects.
To avoid the same mistakes again, the grounded parameters of the failed action may be saved with the lifted action in order to avoid this grounding again, such that failures as discussed later in Sections~\ref{sec:exp}.A.g)~and~\ref{sec:exp}.A.h) are \textit{learned} and not repeated.
In future work, the real source of the failed execution needs to be determined and different forms of feedback to the learning and planning incorporated, but this is out of scope for this work.

\section{Experiments}
\label{sec:exp}
To evaluate the effect of the proposed algorithms for type-generalized actions, we conducted experiments in a simulated kitchen environment, see Figure~\ref{fig:kitchen}.
In this setup, we focus on the symbolic learning and planning aspects of TAMP, assuming a deterministic environment and known controllers for executing the symbolic actions.
In the kitchen environment, the agent can interact with various different objects including food and tools to manipulate those (see Figure~\ref{fig:hierarchy} for the considered objects and their hierarchy).
The agent can move in four directions and interact with the object in front of it.
These interactions consists of three actions -- \texttt{PICK} to pick up an object, \texttt{PLACE} to place down a hold object, and \texttt{USE} to interact with an object.

Learning is done from observations of some basic tasks executed in the simulated kitchen environment.
The low-level state $x$ coming from the simulator comprises a list of the entities (entities include objects as well as agents) in the environment with observable attributes and relations.
To compute the symbolic state $s$, a set of logical predicates is used to describe relations among one or more entities, which are automatically generated from the observed attributes and used to parse $x$ into $s$, i.e., $s = \Call{Parse}{x}$, such that $s$ consists of all predicates that hold in $x$.
Predicates can be grounded or lifted, where grounded means the predicate has concrete entities as parameters, and lifted means the predicate's parameters are variables.
An action is also referred to as grounded or lifted similarly.

All tasks (during training and evaluation) consist of preparing different food by manipulating food entities with certain tools and the prepared foods have to be served on one plate.
Tasks are specified only with the desired state of the food.
Importantly, only one observation per task is shown -- hence, learning is done from few data and without negative examples.
For training, the following $8$ tasks are demonstrated by a human agent,
\begin{itemize}
	\item \textit{cutting a lettuce}
	\item \textit{cutting a banana}
	\item \textit{blending a carrot}
	\item \textit{blending two carrots}
	\item \textit{toasting a bread}
	\item \textit{toasting two breads}
	\item \textit{cutting a tomato and an onion}
	\item \textit{cutting a lettuce and blending a carrot}
\end{itemize}
which results in $68$ state transitions for learning ($(s,a_d,s')$ tuples), forming the training data $\mathcal{D}$.

In addition to the observed training tasks, the system is given a hierarchy of entities (Figure~\ref{fig:hierarchy}).
This hierarchy is used to learn and imagine generalized actions and for grounding the lifted actions during planning, i.e., to know which concrete objects in the environment can be grounded for specific lifted parameter types.

\plotovercooked

\subsection{Testing learned actions on different task sets}
\label{sec:resultsTaskSets}
To evaluate the generalized actions, we tested them on a variety of tasks requiring different kinds of generalization per task set.
Tasks are divided into six sets of tasks -- S1 to S6 -- and the results are summarized in Table~\ref{tab:planning_results}.
Experiments were done on a standard desktop computer with an Intel i5 2.8GHz processor and 16GB RAM.

All tasks are defined by the grounded predicates that must be fulfilled, e.g., for the task of chopping a \textit{cucumber} \texttt{chop\_state\_Chopped(cucumber\_1:cucumber)}.
A task can involve multiple entities, and all involved entities must be served on one \textit{plate}.
Tasks are considered successfully solved if all goal predicates are fulfilled in a state when the agent executed the plan.

In Table~\ref{tab:planning_results} the evaluation results are summarized.
We compare four variants:
\begin{itemize}
	\item[(1)] \textit{generalized+imagined} using both proposed Algorithms~\ref{alg:learning} and~\ref{alg:imagine},
	\item[(2)] \textit{generalized} using only generalization from demonstrations, i.e, only Algorithm~\ref{alg:learning}, 
	\item[(3)] \textit{individual+imagined} using only the imagination Algorithm~\ref{alg:imagine} with the \textit{individual} actions learned by 
	\item[(4)] the \textit{individual} baseline, based on~\cite{Silver2021}.
\end{itemize}

\paragraph{\textbf{S1} -- demonstrated tasks and objects}
This task set evaluates the replication of demonstrated behavior and consists of tasks from the demonstration data with varied entity configurations.
The evaluations show that for solving already observed tasks with the same objects -- using the learned actions on the same tasks as they were learned from -- the individual actions outperform the generalized versions in terms of planning time.
This can be explained by the fact that the individual actions are more specific and hence there are less \textit{grounded} actions for the planner to consider (although there are way more learned \textit{lifted} individual actions $39$ compared to $13$ generalized actions).

\paragraph{\textbf{S2} -- novel task combinations with demonstrated objects}
With known (demonstrated) objects but novel task combinations, i.e., novel combinations of demonstrated tasks, the individual actions start to fail in some tasks, whereas the generalized actions manage to solve all tasks, i.e., can transfer to novel goal combinations.
The tasks still solved by the individual skills, produce longer (less efficient) solutions (see plan lengths $|p|$).
The imagination algorithm does not help the individual baseline here, as the failures come from task combinations rather than from the involved entity types.

\paragraph{\textbf{S3} -- longer task combinations with demonstrated objects}
When increasing the task complexity in terms of number of involved objects (but only demonstrated objects) and, hence, the length of required plans, the individual actions fail to solve the tasks.
The generalized actions on the other hand are able to transfer to more complex (longer) tasks (see $|p|$, where the maximum observed was $|p| = 15$).
That is made possible through the generalization of the parameter types and the update of the preconditions of the generalized actions, enabling them to cover unseen combinations as well.

\plothierarchy

\plotresulttable

\paragraph{\textbf{S4} -- novel task combinations with novel objects}
In this task set, the transfer to novel task combinations (as in S2) as well as to novel objects is tested.
In order to transfer to novel objects that go beyond the generalizations learnable from demonstrations, the imagination algorithm is required, as it can be seen from Table~\ref{tab:planning_results}.
For example, when only one object type is demonstrated for an action, there is no generalization learnable with Algorithm~\ref{alg:imagine} as it requires at least two similar observations.
Another example, the learned actions generalize to the lowest common ancestor of the observed objects, and, hence, to transfer to an object in a different subtree in the hierarchy, imagination is required as well.
But as these imagined actions are not based on observed data, they may not work when executing in a found plan (or are rejected/confirmed/corrected by a human when proposed by a robot).
Therefore, multiple plan proposals may be required to solve a task, as indicated by the number of plan proposals $\#p$ in the table.

\paragraph{\textbf{S5} -- novel task combinations with novel objects from demonstrated classes}
While the learned and imagined generalized actions may produce similar actions and solve similar tasks, the learned actions are backed up by training data, i.e., successful observations.
The imagined actions are \textit{proposals} that may be used to solve the otherwise unsolvable task -- but the imagination algorithm allows the system to propose such \textit{explorative} actions and plans that can be tested by the agent.
To evaluate the different behavior of learned and imagined generalized actions, this task set tests novel task combinations and objects (as S4), but restricts the novel objects to objects from the same classes as the demonstrated ones.
As we can see from the result, both generalized actions can solve the presented tasks.
But comparing the planning time and number of required plan proposals, the learned actions outperform the imagined actions.
The imagined actions are created during planning time, and, hence, it is more costly compared to using the previously learned generalized actions.
Additionally, as discussed in S4 before, imagined actions may require multiple plan proposals, increasing the planning costs as well.
Thus, storing the generalized knowledge in learned generalized actions is more efficient compared to \textit{on-the-fly} generalization through imagination -- but imagination provides explorative actions and plans to potentially solve more tasks and gather additional information (training data) to update the learned actions.

\paragraph{\textbf{S6} -- longer task combination with novel objects}
In this task set, all previous generalization cases are combined, requiring the transfer to unrestricted novel objects as well as to longer task combinations.
Here, both algorithms are required to solve the presented tasks, showing the strength of combining the two complementary approaches.
Learning generalized actions provides condensed and transferable knowledge from the demonstrated data, and as imagination is using the learned actions as starting point, these learned actions provide a broader foundation for more expressive imagined action proposals.

\paragraph{Simulating broken objects}
Additionally, we tested our approach in environments that behave unexpectedly, i.e., objects do not behave as demonstrated, for example, due to a malfunction in a tool.
As this knowledge is not available during planning, the proposed plan tries to use all the objects as learned.
During execution, the unexpected behavior is observed and a new plan is required.
Due to the imagination algorithm, a novel tool (with a similar effect) can be directly proposed in a new plan, and the agent can either try to execute it to test it, or wait for confirmation by a human if the proposed action is acceptable or safe.
Our approach was able to propose successful plans in these scenarios.

\paragraph{Testing exception cases}
Similar, in another evaluation setup, we test exception cases. 
For example, the learned \textit{cutboard} action has generalized to the lowest common ancestor \textit{food} from the demonstrations.
But such generalizations may cover exception cases that are not true, for example, a \textit{coconut} may not be cuttable by the \textit{cutboard} tool, even though belonging to the \textit{food} class.
After observing the failure during execution of the first plan (or again by feedback from a human), the approach is able to propose an alternative in the next plan, here for example, using a \textit{machete} instead.
The new successful execution can be used to update the learned action, whereas the failed execution is stored to avoid the same mistake again.
Our approach was able to adapt to these scenarios and propose successful plans.

\subsection{Evaluation of number of demonstrations}
\label{sec:resultsSplits}
To evaluate the effect of the given amount of demonstrations, we tested our approach with learning from different subsets of the $8$ demonstrations.
The results are summarized in Figure~\ref{fig:result_summary}, showing the median and distribution of successfully solved tasks.
For each tested demonstration subset configuration, we learned action sets for all demonstration combinations -- $8\choose k$ with $2 \leq k \leq 8$ -- and tested each on the tasks from all task sets (S1 -- S6).
In general and as expected, demonstrating more tasks increases the median performance as well as the upper and lower bounds. 
Surprisingly, there is big jump in performance when going from $4$ to $5$ demonstrations, with some combinations already solving all tasks.
As success is not only dependent on the demonstrations alone, but rather on the combination of demonstrated and tested tasks, this indicates an interesting next step involving adding active learning, to \textit{ask} for the most helpful next demonstration in a human-robot-interaction scenario.

\plotresultsplit

\subsection{Discussion \& Outlook}
The proposed approach for learning (and imagine) type-generalized actions can learn from few positive observation and quickly propose additional actions during planning.
With a diverse set of evaluation tasks, the proposed approach showed its transferability to novel entities and problems, being able to solve all tasks.
In contrast to the individual actions baseline, which is only able to solve tasks from or close to the training data.
However, the proposed approach has two main assumptions and limitations: 

(1) the environment observation ($x$) needs to include the type of the entity as well as the attributes describing the entity (and possible relations among those), as this information is used to automatically generate the symbolic state $s$ consisting of logical predicates.
We focused on learning the symbolic actions given such a symbolic state representation, the learned actions are only as good as the state representation. 
Especially in real-world setting like robotics, this grounding problem -- getting symbolic entity information from sensor data -- is a crucial and difficult step~\cite{Roy2014,Taniguchi2019,roy2021machine,kroemer2021review}, but out of scope for this work.

(2) the entity hierarchy is assumed to be known for type-generalization.
We think that this is not a strong assumption, since it has been shown how such hierarchies can be extracted or learned from diverse knowledge sources and common knowledge~\cite{University2010, Beetz2018,Eggert2020}, or (additionally) learned through (virtual) human-robot interaction~\cite{venkatesh2020teaching,deigmoller2022situational,wang2023explainable}.

The hierarchy could also be learned or updated alongside the actions by adding new abstract entities based on similar behavior of entities and using similarity measurements to link novel entities to the hierarchy.

In future work, interactive and interleaved learning with the planner will be investigated in order to make use of the (failure) feedback of executed plans.
Updating the learned actions with successfully executed proposed imagined actions improves performance by adding this knowledge, hence, no repeated imagination is required.
Failed executions on the other hand, no matter if from learned or imagined actions, provide valuable information to update the learned actions, especially to improve the preconditions.

\section{Conclusion}
We propose an approach consisting of two algorithms that can create type-generalized symbolic actions from few observations and a given entity hierarchy.
These actions can be used by a heuristic planner to solve novel task combinations involving novel entities, leveraging the learned type-generalizations and proposing new generalizations during planning if required.
In a simulated kitchen environment, the approach successfully showed the benefits of such type-generalized actions for transferability and set the foundation for future work on applying it to more complex robotic tasks and incremental lifelong learning.

\addtolength{\textheight}{-4cm}   




\bibliographystyle{IEEETran.bst}
\bibliography{learning_hierarchical}

\end{document}